\documentclass{article}

\usepackage{microtype}
\usepackage{graphicx}
\usepackage{subfig}
\usepackage{float}
\usepackage{booktabs} 
\usepackage{tikz}
\usetikzlibrary{calc}
\usepackage{relsize}
\tikzset{fontscale/.style = {font=\small\relsize{#1}}
    }
\usepackage{amsmath}
\usetikzlibrary{shapes.geometric, arrows, backgrounds, fit}
\tikzstyle{layer} = [rectangle, rounded corners, minimum width=1cm, minimum height=0.5cm,text centered, draw=black, fill=red!30]
\tikzstyle{cell} = [rectangle, minimum width=1cm, minimum height=0.5cm,text centered, draw=black, fill=green!30]
\tikzstyle{state} = [rectangle, minimum width=1cm, minimum height=0.5cm,text centered, draw=black, fill=blue!30]
\tikzstyle{arrow} = [thick,->,>=stealth]

\usepackage{hyperref} 

\usepackage[accepted]{icml2019}

\begin{document}

\twocolumn[
\icmltitle{Multi-Objective Reinforced Evolution in Mobile Neural Architecture Search}



\icmlsetsymbol{equal}{*}

\begin{icmlauthorlist}
\icmlauthor{Xiangxiang Chu}{equal,mi}
\icmlauthor{Bo Zhang}{equal,mi}
\icmlauthor{Ruijun Xu}{equal,mi}
\icmlauthor{Hailong Ma}{equal,mi}

\end{icmlauthorlist}

\icmlaffiliation{mi}{Xiaomi AI, Beijing, China}

\icmlcorrespondingauthor{Xiangxiang Chu}{chuxiangxiang@xiaomi.com}
\icmlcorrespondingauthor{Bo Zhang}{zhangbo11@xiaomi.com}

\icmlkeywords{Deep Learning, Neural Architecture Search, ICML}

\vskip 0.3in
]



\printAffiliationsAndNotice{}  

\begin{abstract}
Fabricating neural models for a wide range of mobile devices demands for a specific
 design of networks due to highly constrained resources. Both evolution algorithms
(EA) and reinforced learning methods (RL) have been dedicated to solve 
neural architecture search problems. However, these combinations usually concentrate on a single 
objective such as the error rate of image classification. They also fail to harness
the very benefits from both sides. In this paper, we present a new multi-objective 
oriented algorithm called MoreMNAS (\textbf{M}ulti-\textbf{O}bjective 
\textbf{R}einforced \textbf{E}volution in \textbf{M}obile \textbf{N}eural 
\textbf{A}rchitecture \textbf{S}earch) by leveraging good virtues from both EA 
and RL. In particular, we incorporate a variant of multi-objective genetic algorithm NSGA-II, 
in which the search space is composed of various cells so that crossovers and 
mutations can be performed at the cell level. Moreover, reinforced control is mixed with a natural mutating process to regulate arbitrary mutation, maintaining a delicate balance
between \textit{exploration} and \textit{exploitation}. Therefore, not only does our method 
prevent the searched models from degrading during the evolution process, but it also makes better
use of learned knowledge. Our  experiments conducted in Super-resolution domain (SR) deliver rivalling models compared to some state-of-the-art methods with  fewer FLOPS \footnote{Our generated models along with their metrics are released at \url{https://github.com/moremnas/MoreMNAS}}. 
\end{abstract}

\section{Introduction}
\label{introduction}

Lately, automated neural architecture search has witnessed a victory versus human experts, confirming itself as the next generation paradigm of architectural engineering. The innovations are exhibited mainly in three parts: search space design, search strategy, and evaluation techniques.

The search space of neural architectures is tailored in various forms with respect to different search strategies. It is represented as a sequence of parameters to describe raw layers in NAS \cite{zoph2016neural}, followed by MetaQNN \cite{baker2017designing}, ENAS \cite{pham2018efficient}, in which the selection of parameters is essentially finding subgraphs in a single directed acyclic graph (DAG). Inspired by successful modular design as in Inception \cite{szegedy2015going}, meta-architectures like stacks of blocks \cite{dong2018ppp-net} or predetermined placement of cells \cite{zoph2017learning,liu2017progressive} become a favourite. Each block or cell is a composition of layers, the search space is then outlined by operations within each module, like altering filter size and number, varying layer type, adding skip connections etc. Fine granular network description down to raw layers is more flexible but less tractable, while the coarse one with modular design is merely the opposite. Whereas in most genetic settings, a genotype representation such as an encoding of binary strings is preferred as in GeneticCNN \cite{Xie2017Genetic}, and in NSGA-Net \cite{lu2018nsga}.

Smart search strategy can avoid abusive search in the vast space of neural architectures. Diverse investigations have been made with both evolutionary algorithms and reinforced learning, though mostly independent. Pure reinforced methods was initiated by NAS \cite{zoph2016neural}, later echoed by NASNet \cite{zoph2017learning}, ENAS \cite{pham2018efficient}, MetaQNN \cite{baker2017designing}, MnasNet \cite{tan2018mnasnet}, MONAS \cite{hsu2018monas} etc. Meanwhile, evolutionary approaches contain GeneticCNN \cite{Xie2017Genetic}, NEMO \cite{Kim2017NEMON}, LEMONADE \cite{elsken2018efficient}, \cite{real2018regularized}, NSGA-Net \cite{lu2018nsga}.  Other searching tactics like sequential model-based optimization (SMBO) are also exhibited in DeepArchitect \cite{negrinho2017deeparchitect}, PNAS \cite{liu2017progressive}, and in PPP-Net \cite{dong2018ppp-net}. Notice that multi-objective solutions exist in all categories.

Nevertheless, the idea of putting RL and EA methods together is alluring since the former might degrade and the latter is sometimes less efficient, RENAS \cite{chen2018reinforced} attempted to mitigate this problem with integration of both methods, but unfortunately, they made the false claim since it still fails to converge.

Typical evaluation involves training generated models and performing validation on a held-out set. Due to its computational cost, these models are normally not fully trained, based on the empirical conception that better models usually win at early stages. Less training data also facilitates the process but introduces biases. Other approaches seek to save time by initializing weights of newly morphed models with trained ones \cite{pham2018efficient}, \cite{elsken2018efficient}.

In this paper, we demonstrate a multi-objective reinforced evolutionary approach MoreMNAS to resolve inherent issues of each approach. Our approach differs from previous works by:
\begin{enumerate}
\item inheriting the advantages from both NSGA-II and reinforcement learning to perform multi-objective NAS while overcoming the drawbacks from each method,
\item  construction of cell-based search space to allow for genetic crossover and mutation, hierarchical organization of reinforced mutator with a natural mutation to assure the convergence and to speed up genetic selection,  
\item  applying multi-objective NAS for the first time in Super-resolution domain other than common classification tasks and the results dominate some of the state-of-the-art deep-learning based SR methods, with the best models aligned near the Pareto front in a single run,
\item involving minimum human expertise in model designing as an early guide, and imposing some practical constraints to obtain feasible solutions.
\end{enumerate}

Experiments show that our generated models dominate some of the state-of-the-art methods: SRCNN \cite{dong2014learning}, FSRCNN \cite{dong2016accelerating}, and VDSR \cite{kim2016accurate}. 

\section{Related Works}

\subsection{Single-Objective Oriented NAS}

The majority of early NAS approaches fall in this category, where the validation accuracy of the model is the sole objective.

\subsubsection{Reinforcement Learning-Based Approaches}
Excessive research works have applied reinforcement learning in neural architecture search as aforementioned. They can be loosely divided into two genres according to the type of RL techniques: Q-learning and Policy Gradient. 

For Q-learning based methods like MetaQNN \cite{baker2017designing}, a learning agent interacts with the environment by choosing CNN architectures from finite search space. It stores validation accuracy and architecture description in replay memory to be periodically sampled. The agent is enabled with $\epsilon$-greedy strategy to balance exploration and exploitation.

In contrast, policy gradient-based methods feature a recurrent neural controller (RNN or LSTM) to generate models, with its parameters updated by a family of Policy Gradient algorithms: REINFORCEMENT \cite{zoph2016neural, pham2018efficient,hsu2018monas}, Proximal Policy Optimization \cite{zoph2017learning,tan2018mnasnet}. The validation accuracy of models is constructed as a reward signal. The controller thus learns from experience and creates better models over time. Although some of them have produced superior models over human-designed ones on pilot tasks like CIFAR-10 and MNIST, these RL methods are subject to convergence problems, especially when scaling is involved. Besides, they are mainly single-objective, limiting its use in practice.

\subsubsection{Evolution-Based Approaches}

Pioneering studies on evolutionary neural architecture search form a subfield called Neuroevolution. For instance, NEAT \cite{stanley2002evolving} has given an in-depth discussion of this field, while itself evolves network topology along with weights to improve efficiency, but till recent works like GeneticCNN \cite{Xie2017Genetic}, \cite{real2018regularized}, and \cite{liu2017hierarchical}, evolutionary approaches become substantially enhanced and practical to apply.

In this class, a population of neural models is initialized either randomly or non-trivially. It propagates itself through \emph{crossover} and \emph{mutation}, or \emph{network morphism}, less competitive models are eliminated while the others continue to evolve. In this way, the reduction of traverses in search space is paramount. The selection of an individual model in each generation is based on its \emph{fitness}, e.g. higher validation accuracy. 

In general, recent EA methods are proved to be on par with their RL counterparts and superior against human artistry, attested by AmeobaNet-A  presented in \cite{real2018regularized}.

\subsubsection{Reinforced Evolution-based Approaches}

An attempt to resolve the gap between EA and RL goes to RENAS \cite{chen2018reinforced}, where a reinforced mutation replaces random mutation in order to avoid degradation. The integration, however, fails to assure convergence in the stability test, because the selected genetic algorithm doesn't preserve advantages among generations. Better models once generated are possibly removed during its evolution process. 

\subsection{Multi-Objective Oriented NAS}

Deploying neural models in a wide range of settings naturally requires a multi-objective perspective. MONAS \cite{hsu2018monas} extends NAS by scheming a linear combination of prediction accuracy and other objectives in this regard. But according to \cite{deb2002fast}, a linear combination of objectives is suboptimal. Hence it is necessary to embed this search problem in a real multi-objective context, where a series of models is found along the \emph{Pareto front} of multi-objectives like accuracy, computational cost or inference time, number of parameters, etc. 

LEMONADE \cite{elsken2018efficient} utilizes Lamarckian inheritance mechanism which generates children with performance warm-started by their parents. It also makes the use of the fact that the performance of models (proxied by the number of parameters, multi-adds) are much easier to calculate than evaluation. 

PPP-Net \cite{dong2018ppp-net} is a multi-objective extension to \cite{liu2017progressive}, where Pareto-optimal models are picked in each generation.  It also introduces a RNN regressor to speed up model evaluation. 

NEMO \cite{Kim2017NEMON} and  NSGA-Net  \cite{lu2018nsga} adopt the classic non-dominated search algorithm (NSGA-II)  to handle trade-off among multi-objectives. It groups models based on dominance, while measuring crowding distance gives priority to models within the same front. Besides, NSGA-Net uses Bayesian Optimization to profit from search history.

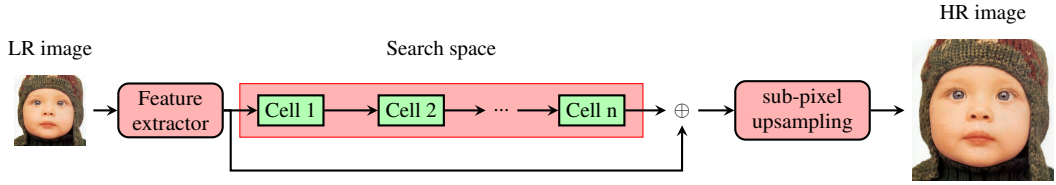
\begin{figure*}[ht]
\vskip 0.2in
\begin{center}
\centerline{
	\begin{tikzpicture}[thick,scale=0.8, every node/.style={scale=0.8},node distance=2cm] %
	\node (lr)  [label={LR image}] {\includegraphics[scale=0.13]{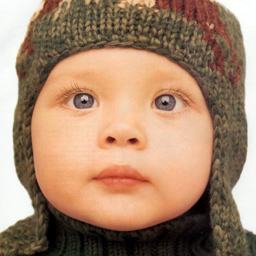}};
	\node (extract) [layer, right of=lr, text width=1.5cm] {Feature extractor};
	\node (cell_0) [cell, right of=extract] {Cell 1};
	\node (cell_1) [cell, right of=cell_0] {Cell 2} ;
	\node (ellipsis) [right of=cell_1,xshift=-0.5cm] {...} ;
	\node (cell_n) [cell,right of=ellipsis,xshift=-0.5cm] {Cell n} ;
	\node (add) [right of=cell_n,xshift=-0.5cm] {$\oplus$};
	\node (upsample) [layer, right of=add, text width=2cm] {sub-pixel upsampling};
	\node (hr) [right of=upsample,label={HR image},xshift=1cm] {\includegraphics[scale=0.13]{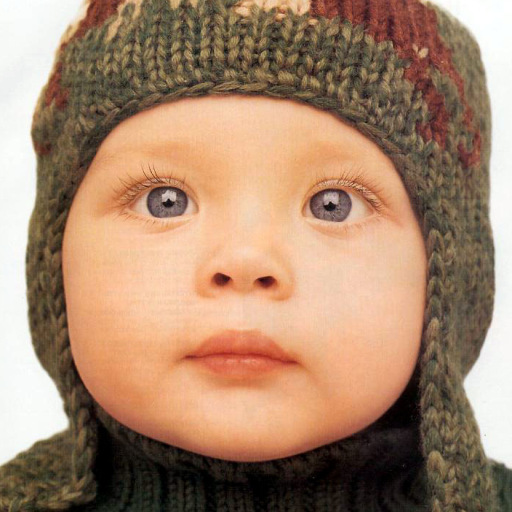}};
	\node [scale=1.3,fit=(cell_0)(cell_1)(ellipsis)(cell_n)] (space) {};
	\begin{pgfonlayer}{background}	
		\filldraw [fit=(space),fill=red!30,draw=red](space.south -| space.west) rectangle (space.north -| space.east);
	\end{pgfonlayer}	
	\node [above of=space, yshift=-1cm] {Search space};
	\draw [arrow] (lr) -- (extract);
	\draw [arrow] (extract) -- (cell_0);
	\draw [arrow] (cell_0) -- (cell_1);
	\draw [arrow] (cell_1) -- (ellipsis);
	\draw [arrow] (ellipsis) -- (cell_n);
	\draw [arrow] (cell_n) -- (add);
	\draw [arrow] (add) -- (upsample);
	\draw [arrow] (upsample) -- (hr);
	\draw [arrow] (extract) -- ++(1,0cm) -- ++(0,-1cm)  -- ++(7.5,0cm) -- (add);
	\end{tikzpicture}
}
\caption{Neural Architecture of Super-Resolution.}
\label{fig:nas-sr}
\end{center}
\vskip -0.2in
\end{figure*}

\section{Multi-Objective Reinforced Evolution}
\subsection{Building Single Image Super-Resolution as a CMOP}

Single image super-resolution (SISR) is a classical low-level task in computer vision. Deep learning methods have obtained impressive results with a large margin than other methods. However, most of the studies concentrate on designing deeper and more complicated networks to boost PSNR (peak-signal-noise-ratio) and SSIM (structural similarity index), both of which are commonly used evaluation metrics \cite{wang2004image,hore2010image}. 

In practice, the applications of SISR are inevitably constricted. For example, mobile devices usually have too limited resources to afford heavy computation of large neural models. Moreover, mobile users are so sensitive to responsiveness that the inference time spent on a forward calculation must be seriously taken into account. As a result, engineers are left with laborious model searching and tuning for devices of different configurations. In fact, It is a multi-objective problem (MOP). Natural thinking that transforms MOP into a single-objective problem by weighted summation is however impractical under such situation. In addition, some practical constraints such as minimal acceptable PSNR must be taken into account.

Here, we rephrase our single image super-resolution as a constrained multi-objective problem as follows,
\begin{equation}\label{eq:sr_mop}
\begin{split}
\min_{m\in S} &\quad  objs(m) \\
s.t. &\quad psnr_- < psnr(m) < psnr_+ \\
  &\quad  flops_- < flops(m) < flops_+ \\
  &\quad  params_- < params(m) < params_+ \\
\text{where} &\quad objs(m) = \{(-psnr(m), flops(m), \\
&\quad \quad params(m)) |  m \in S \}.
\end{split}
\end{equation}
In Equation~\ref{eq:sr_mop}, $m$ is a model from the search space $S$, and our objectives are to maximize the PSNR of a model evaluated on benchmarks while minimizing its FLOPS and number of parameters. The $_-$ subscript represents a lower bound and $_+$ means the corresponding upper constraint. In practice, $flops_-$ and $params_-$ can be set as zero to loosen the left side constraints while the $psnr_+$ can be set as $+\infty$ for the right side.  Here, the model architecture is the \emph{decision variable}. Generally speaking, the above objectives are competing or even conflicting. As a consequence, no model can achieve an optimum across all objectives. 

\subsection{Architecture}

\begin{figure}[ht]
\begin{center}
\centerline{
	\begin{tikzpicture}[thick,scale=0.8, every node/.style={scale=0.8},node distance=2cm]
	\node (cell_0) [cell] {Cell 1};
	\node (cell_1) [cell, below of=cell_0,yshift=1cm] {Cell 2};
	\node (ellipsis)[below of=cell_1,yshift=1cm] {...};
	\node (cell_n) [cell, below of=ellipsis,yshift=1cm] {Cell n};
	\node [scale=1.2,fit=(cell_0)(cell_1)(ellipsis)(cell_n)](space){};
	\node [above of=space, yshift=0.2cm] {Search Space};
	\begin{pgfonlayer}{background}	
		\filldraw [fit=(space),fill=red!30,draw=red](space.south -| space.west) rectangle (space.north -| space.east);
	\end{pgfonlayer}	
	\node (crossover) [state,right of=space, xshift=1cm, yshift=0.5cm] {Cell-based Crossover};
	\node (mutation) [state, below of=crossover,yshift=1cm] {Hierarchical Mutation};
	\node [scale=1.2,fit=(crossover)(mutation)] (controller) {};
	\node [above of=controller, yshift=-0.5cm] {NSGA-II Controller};
	\begin{pgfonlayer}{background}	
		\filldraw [fit=(controller),fill=red!30,draw=red](controller.south -| controller.west) rectangle (controller.north -| controller.east);
	\end{pgfonlayer}	
	\node (evaluator) [layer, xshift=2cm,right of=controller] {Evaluator};
	\draw [arrow] (space) -- (controller);
	\draw [<->,>=stealth,thick] (controller) -- (evaluator) node[midway,above]{train} node[midway, below,text width=1cm]{multi-reward};
	\end{tikzpicture}
}
\vskip 0.1in
\caption{The MoreMNAS Architecture.}
\label{fig:moremnas-arch}
\end{center}
\vskip -0.2in
\end{figure}
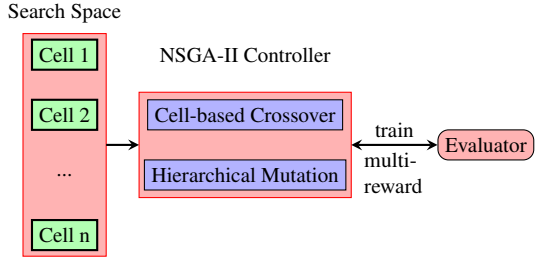

As shown in Figure~\ref{fig:moremnas-arch}, MoreMNAS contains three basic components: cell-based search space, a model generation \textbf{controller} regarding multi-objective based on NSGA-II and an \textbf{evaluator} to return multi-reward feedback for each child model. In our pipeline, the controller samples models from the search space, and it dispatches them to the evaluator to measure their performances, producing feedback for the controller in return.

In specific, the search space is composed of $n$ cells, and each cell contains one or more basic operator blocks. Each block within a cell has $m$ repeated basic operators so that it can represent various architectures. Moreover, all cells can be different from others. For simplicity, our cells share the same search space of repeated blocks.

As for the controller, we use an NSGA-II variation similar to NSGA-Net \cite{lu2018nsga}. Unlike NSGA-Net, we perform crossovers on a cell-based architecture other than on binary encodings. Furthermore, our approach differs from NSGA-Net by a two-level hierarchical mutation process, which balances the trade-off between exploration and exploitation.

\subsection{Search Space}
Proper search space design is of great benefit to boost final performance. In fact, macro-level search \cite{zoph2016neural} is harmful to mobile devices regarding some underlying hardware designs \cite{ma2018shufflenet}. It suffers too much from the non-regularized combinations of arbitrary basic operators. Therefore, quite a few recent research no longer searches on the basis of wild basic operators and makes use of some good cells already discovered \cite{zoph2017learning, tan2018mnasnet}. Partly motivated by MNAS \cite{tan2018mnasnet}, we also construct neural models on top of various cells. Unlike classification tasks, a super-resolution process can be divided into three consecutive sub-procedures: feature extraction, non-linear mapping, and reconstruction \cite{dong2014learning}. Most of the hottest research focuses on non-linear mapping process \cite{dong2014learning,lim2017enhanced,ahn2018fast,haris2018deep} where the final performance matters most. Therefore, we build our search space following this procedure. To be specific, we construct neural networks out of a variable backbone with a fixed head and tail: feature extraction and  reconstruction. We make changes only on the central backbone part, where our search space is designed. Any connection topology such as \emph{skip} and \emph{dense} can be introduced as a bonus.

The search space is composed of $n_p$ cells, and each cell contains an amount of repeated basic operators. Different from \cite{zoph2017learning}, we don't place the same structure constraint for cells in one model. Hence, the operations for a single model can be represented by $s=(s_1,..., s_{n})$, where $s_{n}$ is an element from the search space $S_{n}$ for $cell_{n}$. Within $S_{i}$, we use the combinations to act as a basic element.
Particularly, $cell_{i}$ contains the following basic operators for super-resolution,
\begin{itemize}
	\item basic convolutions: 2D convolution, inverted bottleneck convolution with an expansion rate of 2, grouped convolution with groups in \{2,4\}
	\item kernel sizes in \{1, 3\}
	\item filter numbers in \{16, 32, 48, 64\}
	\item whether or not to use skip connections
	\item the number of repeated blocks in \{1, 2, 4\}
\end{itemize}

For example, two repeated $3\times3$ 2D convolutions with 16 channels is a basic operator. By this mean, each basic element can be presented by a unique index. The total search domain has the following complexity:
$c(s)=\prod_{i=1}^{n}c(s_i)$. If we choose the same operator set with $c$ elements for each cell, then the complexity can be simplified as $c(s)=c^{n}$, where $c=4\times2\times4\times2\times3=192$ and $n = 7$ in our
experiments.
Moreover, this approach constructs a one-to-one mapping between the cell code and neural architectures, possessing an inherent advantage over NSGA-Net, whose space design involves a many-to-one mapping that further repetition removal steps must be taken to avoid meaningless training labors \cite{lu2018nsga} \footnote{In fact, there are some many-to-one mappings within this design. However, their low probability makes them ignorable.}.

Unlike classification task whose output distribution is often represented by a softmax function, single image super-resolution transforms a low-resolution picture to a high-resolution target. Its search space is composed of operators different from those of classification. If pooling and batch normalization are included, super-resolution results will deteriorate \cite{lim2017enhanced}. Besides, a down-sampling operation is excluded since it only blurs the features irreversibly.

\subsection{Model Generation based on NSGA-II with Cell-level Crossover and Hierarchical Mutation}

In this section, we only state those changed components of the original NSGA-II algorithm and the rest is skimmed.
\subsubsection{Population Initialization}
A good initialization strategy usually involves good diversities, which is beneficial to speed up the evaluation. In this paper, we initialize each individual by randomly selecting basic operators for each cell from search space and repeating
the cell for $n$ times to build a model. We call this process \emph{uniform initialization}. In such a way, we generate $2N$ individuals in total for the first generation.
\subsubsection{Non-dominated Sorting}
We use the same strategy as NSGA-II with an improved crowding distance to address the existing problem of original definition that doesn't differentiate individuals within the same cuboid \cite{chu2018improved}.

The improved crowding distance for individual $j$ in $n$-th order for objective $k$ is defined as follows,
\begin{equation}
	dis^j = \sum_{k=1}^{K} \frac{f_{n+1}^k-f_n^k}{f_{max}^k-f_{min}^k}
\end{equation}
where $f^k_n$ is a fitness value for an individual in $n$-th order sorted by objective $k$.

\subsubsection{Cell-based Crossover}
Unlike the encoding model with 0-1 bits in NSGA-Net, we take a more natural approach to perform crossover as the model consists of various cells by design. A crossover of two individuals (i.e., models) $x=(x^1,...,x^i,...,x^{n_p})$ and $y=(y^1,...,y^i...,y^{n_p})$ can result in a new child  $z=(x^1,...,x^{i-1},y^i,x^{i+1},...,x^{n_p})$ when a single-point crossover is performed at position $i$. Other strategies such as \emph{two-point} and \emph{$k$-point} crossovers can also be applied \cite{holland1975adaptation,gwiazda2006crossover,yu2010introduction}.
\subsubsection{Hierarchical Mutation}
While a crossover mainly contributes to exploitation, a mutation is usually aimed to introduce exploration. Some evolution-based NAS methods only contain reinforced mutation to balance exploitation and exploration \cite{chen2018reinforced,hsu2018monas}. Therefore, their result heavily depends on the reinforced mutation, which is sensitive to hyper-parameter tuning and reward design. Since the total pipeline of neural architecture search is time-consuming, it's difficult and even intractable to make a satisfactory setup by trial and error. On the other hand, pure evolution based NAS algorithms seldom makes good use of the knowledge from training the generated models, which can be utilized to guide further search \cite{real2018regularized,cheng2018searching}. To address these problems, we propose hierarchical mutation.

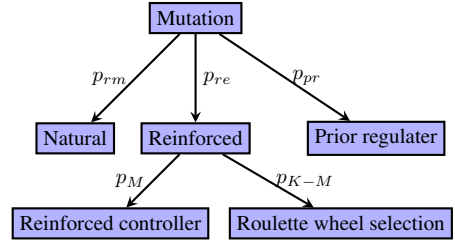
\begin{figure}[ht]
\vskip 0.2in
\begin{center}
\centerline{
	\begin{tikzpicture}[thick,scale=0.8, every node/.style={scale=0.8},node distance=2cm]
	\node (mutation) [state] {Mutation};
	\node (re) [state,below of=mutation] {Reinforced};
	\node (natural) [state,left of=re] {Natural};
	\node (prior) [state, right of=re, xshift=1cm] {Prior regulater};
	\node (re-con) [state, below left of=re] {Reinforced controller};
	\node (roulette) [state, below right of=re, xshift=1cm] {Roulette wheel selection};
	\draw [arrow] (mutation) -- (re) node[midway,right] {$p_{re}$} ;
	\draw [arrow] (mutation) -- (natural) node[midway,left] {$p_{rm}$} ;
	\draw [arrow] (mutation) -- (prior) node[midway,right] {$p_{pr}$} ;
	\draw [arrow] (re) -- (re-con) node[midway, left] {$p_M$};
	\draw [arrow] (re) -- (roulette) node[midway, right] {$p_{K-M}$};
	\end{tikzpicture}
}
\vskip 0.1in
\caption{Hierarchical mutation.}
\label{fig:mixed-mutator}
\end{center}
\vskip -0.4in
\end{figure}

The top level of hierarchical mutation includes three mutually exclusive processes: \emph{reinforced dominant}, \emph{natural} and \emph{prior-regularized} mutation. Whenever a new individual is prepared to be mutated, we sample from a category distribution over these three mutations, i.e. $p(mutation)=\{p_{re}, p_{rm}, p_{pr}\}$.

The word reinforced dominant means that we use reinforcement learning to minimize objectives which are otherwise hard to predict. Actually, almost all objectives of multi-objective problems can be classified into two categories: those difficult to predict and the others not. Without loss of generality, we define $K$ objectives, the first $M$ ones are difficult to predict, and the remaining ones are not. 

In super-resolution domain, for instance, the \emph{mean square error} (MSE) is hard to obtain since it is computed between a ground truth high-resolution picture and the one generated by a deep neural model. In our optimization problem defined by Equation~\ref{eq:sr_mop}, we have $K=3$ objectives in total, where $M=1$ objective is $-psnr$, and $K-M=2$ objectives are respectively \emph{multi-adds} and \emph{number of parameters}. 

We take different steps to handle these two categories. For the former, we use $M$ reinforced controllers, one for each objective. For the latter, we use \emph{Roulette-wheel selection} to sample one cell in each step \cite{lipowski2012roulette}. 

\begin{figure}[ht]
	\vskip 0.2in
	\begin{center}
	\centerline{
		\vspace*{-45pt} 
		\begin{tikzpicture}[thick,scale=0.8, every node/.style={scale=0.8},node distance=2cm]
			\node (cell_1) [cell] {Cell 1};
			\node (sample_1) [layer, below of=cell_1,yshift=1cm] {sample};
			\node (softmax_1) [layer, below of=sample_1,yshift=1cm] {Softmax};
			\node (lstm_1) [layer, below of=softmax_1,yshift=1cm] {LSTM};
			\node (embed_1) [layer, below of=lstm_1,yshift=1cm] {embedding};
			\node (zero_1) [cell, below of=embed_1,yshift=1cm,fill=green!30] {zero cell};
			\node (start) [state, left of=lstm_1, xshift=0.5cm,text width=1cm] {zero state};
			\node (cell_2) [cell, right of=cell_1,xshift=0.5cm] {Cell 2};
			\node (sample_2) [layer, below of=cell_2,yshift=1cm] {sample};
			\node (softmax_2) [layer, below of=sample_2,yshift=1cm] {Softmax};
			\node (lstm_2) [layer, below of=softmax_2,yshift=1cm] {LSTM};
			\node (embed_2) [layer, below of=lstm_2,yshift=1cm] {embedding};
			\node (ellipsis) [right of=lstm_2,xshift=-0.5cm] {...};
			\node (cell_n) [cell, right of=cell_2,xshift=1cm] {Cell $n$};
			\node (sample_3) [layer, below of=cell_n,yshift=1cm] {sample};
			\node (softmax_3) [layer, below of=sample_3,yshift=1cm] {Softmax};
			\node (lstm_3) [layer, below of=softmax_3,yshift=1cm] {LSTM};
			\node (embed_3) [layer, below of=lstm_3,yshift=1cm] {embedding};
			\draw [arrow] (zero_1) -> (embed_1);
			\draw [arrow] (embed_1) -> (lstm_1);
			\draw [arrow] (lstm_1) -> (softmax_1);
			\draw [arrow] (softmax_1) -> (sample_1);
			\draw [arrow] (sample_1) -> (cell_1);
			\draw [arrow] (sample_1.north)[out=90] .. controls +(2,2) and +(-2,-3) .. (embed_2.south);
			\draw [arrow] (embed_2) -> (lstm_2);
			\draw [arrow] (lstm_2) -> (softmax_2);
			\draw [arrow] (softmax_2) -> (sample_2);
			\draw [arrow] (sample_2) -> (cell_2);
			\draw [arrow,dashed] (sample_2.north) [out=90] .. controls +(2,2) and +(-2,-3) .. (embed_3.south);
			\draw [arrow] (embed_3) -> (lstm_3);
			\draw [arrow] (lstm_3) -> (softmax_3);
			\draw [arrow] (softmax_3) -> (sample_3);
			\draw [arrow] (sample_3) -- (cell_n);
			\draw [arrow] (start) -> (lstm_1);
			\draw [arrow] (lstm_1) -> (lstm_2);
			\draw [arrow] (lstm_2)-> (ellipsis);
			\draw [arrow] (ellipsis)-> (lstm_3);
		\end{tikzpicture}
		}
		\vskip 0.1in
		\caption{Controller network for reinforced mutation.}
		\label{fig:rl_controlled_mutation}
	\end{center}
	\vskip -0.2in
\end{figure}
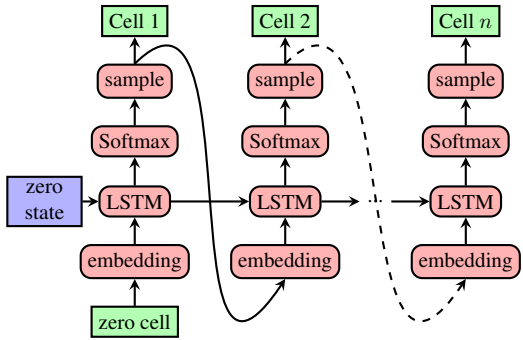

Specifically,  we take advantage of learned knowledge from training generated models and their scores of objectives based on a reinforced mutation controller, which is designed to distil meaningful experience to evaluate model performance. This part of mutation can then be regarded as exploitation. In this procedure, the model generation is described as \emph{Markov Decision Process} (MDP), which is represented by a LSTM controller \cite{hochreiter1997long} shown in Figure~\ref{fig:rl_controlled_mutation}. Initially, the zero state is reset for the LSTM controller, and a cell with index zero is injected into the LSTM controller after embedding. Here the zero index refers to a \emph{null} element which is not related to any basic operator in $S$. Then the controller samples an action from $S$ with a softmax function, i.e. category distribution. In turn, this sampled action serves as an input for the next step. This process is repeated until $n$ cells are generated to build a complete model. 

From the perspective of reinforcement learning, each episode contains $n$ steps. Note that the reward is delayed since it cannot be returned timely until the last cell is generated. In fact, PSNR is not suitable for a direct reward since it is measured in the logarithmic space. Instead, we define the reward based on the mean squared error between a generated high-resolution picture and the ground truth. In particular, the reward of a model $m$ is defined as
\begin{equation}
\label{eq:reward}
	reward(m) = \frac{0.001}{mse_m}-0.5
\end{equation}
The minimization operator is a type of clipping trick to avoid instability from getting a too large value as a reward. Since the reward is obtained after model $m$ is evaluated, which ends an episode, we adopt REINFORCEMENT to update this controller \cite{sutton2018reinforcement}.

Under the assumption of MDP, given a  policy $\pi_\theta$ represented by a neural network parameterized by $\theta$,  its gradient can be estimated in the form of mini-batch $B$ by,
\begin{equation}
\begin{split}
	\nabla_\theta \pi_\theta &=\frac{1}{B}\sum_{b=1}^{B}\sum_{t=1}^{n} r_{t(b)} \log p_\theta(a_{t(b)}|s_{t(b)}) \\
	&=\frac{1}{B}\sum_{b=1}^{B}\sum_{t=1}^{n} \gamma^t r_b \log p_\theta(a_{t(b)}|s_{t(b)})
\end{split}
\end{equation}
where $r_b$ is the reward defined by Equation~\ref{eq:reward} for $b$-th model and $\gamma$ is a discount coefficient.
For simplicity, each cell has the same space configuration $S$.
  
\vspace*{-30pt} 

As for multi-adds and number of parameters, we measure the fitness at logarithmic scale before Roulette-wheel selection. After a rough analysis of the distribution over parameters in a cell search space, the operator with maximum parameters has 1000 times more than the minimum one, which amounts to high probability advantage.

%

Furthermore, natural mutation is applied to encourage exploration throughout the whole evolution. Since \emph{elitist preservation mechanism} can afford no degradation during the model generation process, there is no need to take measures to weaken this natural mutation in the process of evolution. It comes with a byproduct of better exploration. Whereas for mutation controllers based on pure reinforcement learning, some strategies such as discouraging exploration gradually along with the training process are indispensable \footnote{Taking DQN \cite{mnih2015human} for example, decreasing the hyper-parameter $\epsilon$ that allows for exploration during the learning process is a commonly used trick.}. 

As for prior regularized mutation, we are partly motivated by the guidelines from ShuffleNet \cite{ma2018shufflenet}, in which some experimental suggestions regarding hardware implementation are proposed to decrease the running time costs, so did we draw ideas from various renowned neural architectures like ResNet \cite{he2016deep} and MobileNet \cite{sandler2018mobilenetv2}. Apart from that, repeating the same operator many times to build a model is also proved advantageous. However, we introduce this prior with a small probability. To be more precise, we randomly choose a target position $i$ for model $x=(x_1,...,x_i,...,x_{n})$, and we mutate it to generate $x_{child}=(x_i,...,x_i,...,x_i)$.

\subsection{Further Discussion}

\subsubsection{Scalability}

This NAS pipeline can further benefit from large scale parallel computing resources. The total cost is in direct proportion to the total number of  generated models. Therefore, our method can be boosted linearly by increasing computing resources.

\subsubsection{Handling Constraints}
In Equation~\ref{eq:sr_mop}, different constraints can be evaluated at different time. After a model-meta is given, its FLOPS and number of parameters are obtained at the same time. However, the PSNR cannot be measured without regression mechanism unless the model is fully trained.  We can also skip the training procedure if the constraints involving either FLOPS or number of parameters are violated.

The constraint for PSNR is essential since it not only helps to generate and to keep feasible models but excludes bad models with too few parameters, which occupy the precious positions at the Pareto front if not abandoned.

\section{Experiments}
\subsection{Setup and Implementation Details}
In our experiment, the whole pipeline contains 200 generations, during each generation 56 individuals get spawned, i.e., there are 11200 models generated in total. Other hyper-parameters is listed in Table~\ref{tab:pipelienhyper}. In addition, all experiments are performed on a single Tesla V100 machine with 8 GPUs and a complete run takes about 7 days. Detailed hyper-parameters are shown in Tabel~\ref{tab:pipelienhyper}.

\begin{table}[ht]
	\caption{Hyper-parameters for the whole pipeline.}
	\label{tab:pipelienhyper}
	\vskip 0.15in
	\begin{center}
		\begin{small}
			\begin{sc}
				\begin{tabular}{lclc}
					\toprule
					Item & value & Item & value \\
					\midrule
					Population N & 56 & Mutation Ratio  & 0.8 \\
					$p_{rm}$ & 0.50 & $p_{re}$   & 0.45 \\
					$p_{pr}$ & 0.05 & $p_{M}$ & 0.75 \\
					$p_{K-M}$ & 0.25 \\
					\bottomrule
					\end{tabular}
					\end{sc}
					\end{small}
					\end{center}
					\vskip -0.1in
					\end{table}

\subsubsection{Base Model}
Both feature extraction and restoration stage in Figure~\ref{fig:nas-sr} are composed of 32 $3\times3$ 2D convolution filters with a unit stride.
\subsubsection{Evaluator}
Each evaluator trains a dispatched model on DIV2K dataset \cite{timofte2017ntire} across 200 epochs with a batch size of 16. In specific, the first 800 high-resolution (HR) pictures are used to construct the whole training set. The HR pictures are randomly cropped to $80\times80$, and then down-sampled to LR by bicubic interpolation. Furthermore, we use random flipping and rotation with $90^{\circ}, 180^{\circ}$ and $ 270^{\circ}$ to perform data augmentation. Moreover, ReLU acts as the default activation function except for the last layer. We use Adam optimizer to train the model with $\beta_1=0.9$ and $\beta_2=0.999$. The initial learning rate is $1\times 10^{-4}$, and decays by half every 100 epochs. In addition, we use $L_1$ loss between the generated HR images and the ground truth to guide the back-propagation.
\subsubsection{Full Training}
We select several models located at our Pareto front for complete training using the same training set as well as data augmentation tricks. The only difference is that we use a larger cropping size ($128 \times 128$) and a longer epoch 4800. In addition, we also use Adam optimizer to work on this task with $\beta_1=0.9$ and $\beta_2=0.999$. In specific, the initial learning rate is $10^{-3}$ for large models and $10^{-4}$ for small ones. The learning rate decays in half every 80000 back-propagations. 
\subsection{Comparison with Human Designed State of the Art Models}
\begin{figure}[ht]
\vskip 0.2in
\centering
\includegraphics[scale=.7]{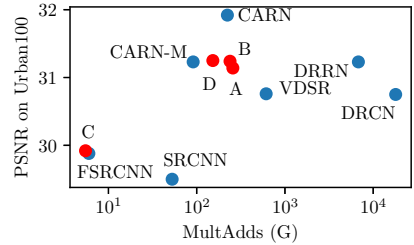}\\
\vskip 0.1in
\caption{MoreMNAS-A,B,C,D (shown in red) vs. others (blue)}
\label{fig:moremnas-vs-sr-urban100}
\vskip -0.2in
\end{figure}

Here we only consider those SR models with comparable parameters and multi-adds. 
In particular, we take SRCNN \cite{dong2014learning}, FSRCNN \cite{dong2016accelerating} and VDSR \cite{kim2016accurate}, and we apply our models at $\times$2 scale since it is one of the most commonly used tasks to draw comparisons. 
Figure~\ref{fig:moremnas-vs-sr-urban100} illustrates our results compared with other state-of-the-art models on Urban100 dataset. 



\begin{figure}[ht]
\vskip 0.2in
\begin{center}
\centerline{
	\begin{tikzpicture}[thick,scale=0.8, every node/.style={scale=0.8},node distance=1cm]
		\node (a) [cell] {invertBotConE2\_f48\_k1\_b4\_noskip};
		\node (b) [cell, below of=a] {invertBotConE2\_f16\_k3\_b4\_noskip};
		\node (c) [cell, below of=b] {invertBotConE2\_f16\_k1\_b4\_isskip};
		\node (d) [cell, below of=c] {groupConG2\_f16\_k1\_b1\_noskip};
		\node (e) [cell, below of=d] {invertBotConE2\_f64\_k3\_b4\_noskip};
		\node (f) [cell, below of=e] {groupConG4\_f64\_k1\_b1\_isskip};
		\node (g) [cell, below of=f] {invertBotConE2\_f48\_k3\_b4\_noskip};
		\draw [arrow] (a) -- (b);
		\draw [arrow] (b) -- (c);
		\draw [arrow] (c) -- (d);
		\draw [arrow] (d) -- (e);
		\draw [arrow] (e) -- (f);
		\draw [arrow] (f) -- (g);
	\end{tikzpicture}
}
\vskip 0.1in
\caption{Model MoreMNAS-A against VDSR}
\label{fig:moremnas-vdsr-55}
\end{center}
\vskip -0.2in
\end{figure}
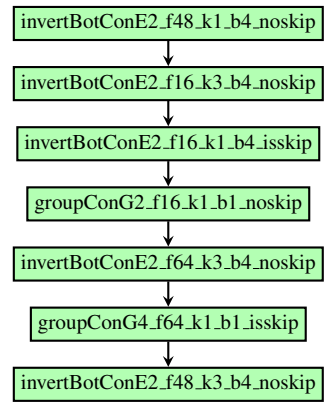

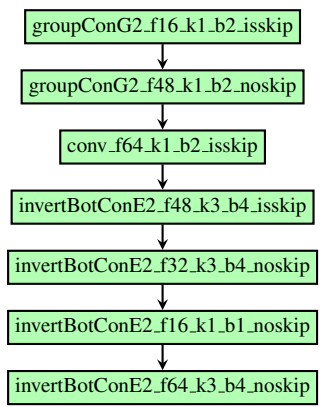
\begin{figure}[ht]
\vskip 0.2in
\begin{center}
\centerline{ 
	\begin{tikzpicture}[thick,scale=0.8, every node/.style={scale=0.8},node distance=1cm]
		\node (a) [cell] {groupConG2\_f16\_k1\_b2\_isskip};
		\node (b) [cell, below of=a] {groupConG2\_f48\_k1\_b2\_noskip	};
		\node (c) [cell, below of=b] {conv\_f64\_k1\_b2\_isskip};
		\node (d) [cell, below of=c] {invertBotConE2\_f48\_k3\_b4\_isskip};
		\node (e) [cell, below of=d] {invertBotConE2\_f32\_k3\_b4\_noskip};
		\node (f) [cell, below of=e] {invertBotConE2\_f16\_k1\_b1\_noskip};
		\node (g) [cell, below of=f] {invertBotConE2\_f64\_k3\_b4\_noskip};
		\draw [arrow] (a) -- (b);
		\draw [arrow] (b) -- (c);
		\draw [arrow] (c) -- (d);
		\draw [arrow] (d) -- (e);
		\draw [arrow] (e) -- (f);
		\draw [arrow] (f) -- (g);
	\end{tikzpicture}
}
\vskip 0.1in
\caption{Model MoreMNAS-B against VDSR}
\label{fig:moremnas-vdsr-95}
\end{center}
\vskip -0.2in
\end{figure}

The detailed results are shown in Table~\ref{tab:psnr_ssim}, where two models we call MOREMNAS-A (Figure~\ref{fig:moremnas-vdsr-55}) and B (Figure~\ref{fig:moremnas-vdsr-95}) hit higher PSNR and SSIM than VDSR across four evaluation sets with much fewer multi-adds. Another model called MOREMNAS-D  dominates VDSR across three objectives with a quarter of its multi-adds. Besides, a light-weight model MOREMNAS-C matches FSRCNN with fewer multi-adds, which again dominates SRCNN across three aspects: higher score, fewer number of parameters and multi-adds. In addition, the non-linear mapping stage contains various cell blocks with a unit kernel, which shares some similarities with FSRCNN.

\begin{table*}[ht]
	\caption{Comparisons with the state-of-the-art methods based on $\times$2 super-resolution task$^\dag$}
	\label{tab:psnr_ssim}
	\vskip 0.1in
	\begin{center}
		\begin{small}
			\begin{sc}
				\begin{tabular}{lrrcccc}
					\toprule
					Model & MultAdds & Params & SET5 & SET14 & B100 & Urban100\\
					& &   &  PSNR/SSIM &  PSNR/SSIM &  PSNR/SSIM &  PSNR/SSIM \\
					\midrule
					SRCNN \cite{dong2014learning} & 52.7G & 57K & 36.66/0.9542 & 32.42/0.9063 & 31.36/0.8879 & 29.50/0.8946 \\ 
					FSRCNN \cite{dong2016accelerating} & 6.0G & 12K & 37.00/0.9558 & 32.63/0.9088 & 31.53/0.8920 & 29.88/0.9020 \\ 
					VDSR \cite{kim2016accurate} & 612.6G & 665K & 37.53/0.9587 & 33.03/0.9124 & 31.90/0.8960 & 30.76/0.9140 \\ 
					
					DRRN \cite{tai2017image} & 6,796.9G & 297K & 37.74/0.9591 & 33.23/0.9136 & 32.05/0.8973 & 31.23/0.9188 \\
					MoreMNAS-A (ours) & 238.6G & 1039K & 37.63/0.9584 & 33.23/0.9138 & 31.95/0.8961 & 31.24/0.9187\\
					MoreMNAS-B (ours) & 256.9G & 1118K & 37.58/0.9584 & 33.22/0.9135 & 31.91/0.8959& 31.14/0.9175\\
					MoreMNAS-C (ours) & 5.5G & 25K & 37.06/0.9561 & 32.75/0.9094& 31.50/0.8904 & 29.92/0.9023\\
					MoreMNAS-D (ours) & 152.4G & 664K & 37.57/0.9584 & 33.25/0.9142 & 31.94/0.8966 & 31.25/0.9191\\
					\bottomrule
					\multicolumn{6}{c} {\footnotesize $^\dag$ \textit{The multi-adds are valued on $480\times480$ input resolution.}}
				\end{tabular}
			\end{sc}
		\end{small}
	\end{center}
	\vskip -0.1in
\end{table*}

It's interesting that \emph{inverted bottleneck} block can also benefit super resolution since it's originally intended for classification and  rarely appears in super-resolution models. Another attractive fact is that \emph{grouped convolution} can also profit super-resolution task. MoreMNAS-A uses inverted bottleneck blocks interleaved with grouped convolutions, while MoreMNAS-B utilizes them separately. Besides, both MoreMNAS-B and D perform very competitively against DRRN \cite{tai2017image} with much fewer multi-adds on Urban100 test set, which is a very challenging task involving rich details.

\begin{figure}[ht]
\begin{center}
\includegraphics{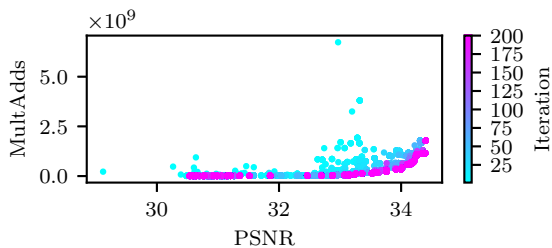}
\vskip -0.1in
\caption{Objectives of the best SR models of each iteration}
\label{fig:pareto-front}
\end{center}
\vskip -0.2in
\end{figure}

More good models generated near the Pareto front are omitted here because of limited space. To sum up, our method can generate various models which dominate those well-known state-of-the-art models with comparable sizes. Figure~\ref{fig:pareto-front} plots the evolutionary process where the latest models push the Pareto front further.

\section{Conclusion}
In this paper, we propose a multi-objective reinforced evolution algorithm in mobile neural architecture search, which seeks a better trade-off among various competing objectives. It has three obvious advantages: no recession of models during the whole process, good exploitation from reinforced mutation, a better balance of different objectives based on NSGA-II and Roulette-wheel selection. To our best knowledge, our work is the first approach to perform multi-objective neural architecture search by combining NSGA-II and reinforcement learning.

Our method is evaluated in the super-resolution domain. We generate several light-weight models that are very competitive, sometimes dominating human expert designed ones such as SRCNN, VDSR across several critical objectives: PSNR, multi-adds, and the number of parameters. Finally, our algorithm can be applied in other situations, not only limited to a mobile setting.

Our future work will focus on the following aspects. First, accelerating the whole pipeline based on a regression evaluation from our sampled data. Second, fine-tuning hyper-parameters and replacing REINFORCEMENT with more powerful policy gradient algorithms such as PPO \cite{schulman2017proximal} and POP3D \cite{chu2018policy}, which have more potential to push the Pareto front further.


%


\nocite{langley00}

\bibliography{nas_nsgaii}
\bibliographystyle{icml2019}



\end{document}